\documentclass[sigconf,screen]{acmart}
\AtBeginDocument{%
  }

\setcopyright{acmlicensed}
\copyrightyear{2018}
\acmYear{2018}
\acmDOI{XXXXXXX.XXXXXXX}
\acmConference[Preprint]{}{Under Review}{2025}
\acmISBN{978-1-4503-XXXX-X/2018/06}

\renewcommand\footnotetextcopyrightpermission[1]{}
\usepackage[utf8]{inputenc}
\usepackage[T1]{fontenc}
\usepackage{pifont}
\usepackage{url}
\usepackage{xcolor}
\usepackage{graphicx}
\usepackage{tabularx}
\usepackage{array}
\usepackage{amsmath}
\usepackage{setspace}
\usepackage{multirow}
\usepackage{makecell}
\usepackage{colortbl}
\usepackage{hhline}
\usepackage{booktabs}
\usepackage{dcolumn}
\usepackage{caption}
\usepackage{subcaption}
\usepackage{chngpage}
\usepackage{enumitem}
\begin{document}

\title{Unveiling the Deficiencies of Pre-trained Text-and-Layout Models in Real-world Visually-rich Document Information Extraction
}


\author{Chong Zhang$^1$, Yixi Zhao$^1$, Yulu Xie$^2$, Chenshu Yuan$^2$, Yi Tu$^3$, Ya Guo$^3$,\\Mingxu Chai$^1$, Ziyu Shen$^1$, Yue Zhang$^1$, Qi Zhang$^1$}
\authornote{Corresponding author.}
\email{{chongzhang20,qz}@fudan.edu.cn}
\affiliation{%
  $^1$School of Computer Science, Fudan University, Shanghai, China \\
  $^2$School of Statistics and Data Science, Nankai University, Tianjin, China \\
  $^3$Ant Tiansuan Security Lab, Ant Group, Hangzhou, China \\
  \country{}
}



\renewcommand{\shortauthors}{Chong Zhang, et al.}

\begin{abstract}
Recently developed pre-trained text-and-layout models (PTLMs) have shown remarkable success in multiple information extraction tasks on visually-rich documents (VrDs).
However, despite achieving extremely high performance on benchmarks, their real-world performance falls short of expectations.
Owing to this issue, we investigate the prevailing evaluation pipeline to reveal that:
(1) The inadequate annotations within benchmark datasets introduce spurious correlations between task inputs and labels, which would lead to overly-optimistic estimation of model performance. 
(2) The evaluation solely relies on the performance on benchmarks and is insufficient to comprehensively explore the capabilities of methods in real-world scenarios.
These problems impede the prevailing evaluation pipeline from reflecting the real-world performance of methods, misleading the design choices of method optimization. 
In this work, we introduce EC-FUNSD, an entity-centric dataset crafted for benchmarking information extraction from visually-rich documents (VrD-IE).
This dataset contains diverse layouts and high-quality annotations. 
Additionally, this dataset disentangles the falsely-coupled segment and entity annotations that arises from the block-level annotation of FUNSD.
Using the proposed dataset, we evaluate the real-world VrD-IE capabilities of PTLMs from multiple aspects, including their absolute performance, as well as generalization, robustness and fairness. 
The results indicate that prevalent PTLMs do not perform as well as anticipated in real-world VrD-IE scenarios. 
We hope that our study can inspire reflection on the directions of PTLM development.

\end{abstract}


\begin{teaserfigure}
  \vspace{0mm}
  \includegraphics[width=\textwidth]{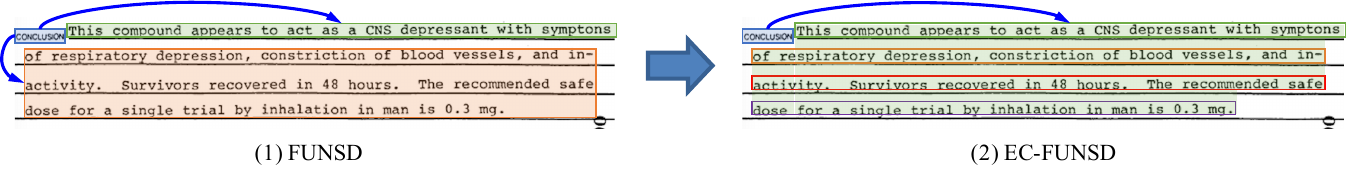}
  \vspace{-8mm}
  \caption{
    A document image with its layout, entity and linking annotations in FUNSD and the proposed EC-FUNSD dataset.
    Each color of shades indicates an annotated entity. 
    (1) Block-level annotations within FUNSD do not correspond to semantic entities, making it unsuitable for entity-centric evaluation. Their linking relationships are also confused.
    (2) On the contrary, EC-FUNSD decouples the layout and entity annotations for proper evaluation.}
  \label{fig:page1}
  \vspace{3mm}
\end{teaserfigure}


\maketitle

\section{Introduction}

The research field of document AI is becoming more popular with the increase in industrial demands \cite{cui2021document,10322990,9897491}. One of the primary objectives in this field is to extract useful information from visually-rich documents (VrDs), given the texts with their xy-coordinates on the document layout. 
In recent years, the advent of pre-trained text-and-layout models (PTLMs) has promoted the understanding of the semantic and spatial relations within the document layouts, and demonstrated great success in multiple visually-rich document information extraction (VrD-IE) tasks\citep{hong2022bros,huang2022layoutlmv3,gu2022xylayoutlm,tu2023layoutmask,cvpr2023geolayoutlm,Liao_2023_ICCV}.
In addressing downstream tasks, the improvement achieved through merely enhancing the downstream task model is limited, due to the scarcity of domain-relevant data resources and the high cost of data annotation. 
In contrast, pre-trained models adapt to the tasks by designing corresponding pre-training tasks.  
These models are able to leverage massive document layout data covering various domains during pre-training to achieve a higher performance.
As a result, it is acknowledged by the document AI community that the recent success of VrD-IE methods is mainly attributed to the advancement in PTLMs, thus the research focus of this field has shifted to the improvement of PTLMs. 
Reliable evaluation of the real-world performance of PTLMs when they are applied in downstream tasks have become essential for the development of document AI.

Conventionally, PTLMs serve as the embedding model for document layouts, just as the role contextualized pre-trained language models played in NLP tasks \cite{kenton2019bert,liu2019roberta,sun2019ernie}. 
Their capabilities for adapting to VrD-IE tasks are usually evaluated by the semantic entity recognition (SER) and entity linking (EL) tasks, where PTLMs typically handle SER in a sequence-labeling manner, and tackle EL with simple classification heads. 
It is widely acknowledged that the requirements for these two tasks are in line with the capabilities required in real-world VrD-IE applications. Therefore, the performance of PTLMs on these two tasks reflects the capacity of their multimodal embeddings to facilitate the VrD-IE tasks.
Currently, state-of-the-art PTLMs have achieved exceedingly good performance on prevailing VrD-IE benchmarks\cite{jaume2019funsd,park2019cord,huang2019icdar2019}. 
As an extreme case, GeoLayoutLM \cite{cvpr2023geolayoutlm} has achieved 100\% test accuracy at the key-value entity linking task of CORD. 
We hope that these seemingly good performance of PTLMs on benchmarks could also be replicated in real-world VrD-IE applications. 

However, the current prevalent benchmarks do not perfectly serve the evaluation of the VrD-IE capabilities of PTLMs, owing to their inherent drawbacks.
For example, FUNSD \cite{jaume2019funsd} and XFUND \cite{xu2022xfund} are the most popular benchmarks in SER evaluation. 
Nevertheless, these datasets were originally designed for visual information extraction tasks and were transformed into the SER format merely for the purpose of evaluation.
The layout semantic annotations within these datasets are in block-level, which emphasizes the spatial relation of text blocks on the vision layouts. 
Some of the blocks do not correspond to semantic entities, making these datasets unsuitable for SER evaluation. 
Besides, 
the other benchmark datasets suffer from (1) the lack of diversity in layout patterns and entity semantics; (2) the insufficiency of complex semantic entities; (3) the granularity-deficiency and low-quality of layout annotations; and (4) the mismatch of task paradigm. 
These issues are further demonstrated in \S\ref{sec:dataset_motivation}, which illustrates the limitation of prevalent datasets for VrD-IE evaluation of PTLMs. 

When these aforementioned unsuitable datasets are utilized in the evaluation pipeline, several potential risks will arise, significantly reducing the reliability of the evaluation.
One of the most serious potential risks is the spurious correlation bias introduced by the block-level layout semantic annotations.
The annotations within several prevailing benchmarks, including FUNSD, XFUND and CORD \cite{park2019cord}, are organized by visual regions (i.e., blocks) of the document layout, in which the semantic category labels and the association relationship between regions are annotated on the block-level.
In SER evaluation, segments and entities are simultaneously represented by these semantic blocks. Therefore, models would predict entity boundaries specified by block annotations with leveraging the segment-level layout inputs from block annotations. Layout elements within the same entity would have exactly the same xy-coordinate inputs, leading to label leakage during training, and resulting in inflated test scores which could not reflect the real model performance. 
Besides, the existing evaluation pipeline is solely centered around the test performance on benchmarks, and fails to comprehensively reflect the real-world performance of methods. 
These issues impede the accurate evaluation of the real-world performance of VrD-IE methods, which may negatively influence the direction of method optimization.

In this paper, our aim is to establish a valid pipeline for evaluating the comprehensive performance of PTLMs in real-world VrD-IE applications. 
The primary step is to introduce a more appropriate benchmark to evaluate the capabilities of PTLMs in VrD-IE tasks. 
Based on the specification of requirements, we propose \textbf{EC-FUNSD}, an \textbf{E}ntity-\textbf{C}entric benchmark derived from FUNSD \citep{jaume2019funsd} that aims to provide a fair and unbiased evaluation for VrD-IE capabilities of PTLMs. 
This benchmark is constructed by manually revising the annotations in FUNSD, and is designed to be used for SER and EL tasks.
Based on the newly proposed unbiased benchmark, we conduct a comprehensive evaluation of prevalent baseline PTLMs on real-world VrD-IE tasks. 
The evaluation aspects include not only the test performance of the model, but also: (1) generalization to unknown distributions; (2) robustness under natural perturbations, and (3) fairness to difficult test sample subsets. 
Experimental results indicate that the true performance of the baseline models is not as good as they have claimed. 
Although EC-FUNSD and FUNSD are very similar in almost all aspects, the baseline models suffer from a significant performance drop on the same task settings, particularly a 7.48-8.55 decrease of F1 on SER. 
This reveals the potential risk of current PTLMs that they may develop to excessively overfit the biased benchmarks, but the actual benefits brought by the advancements are suspicious. 
Moreover, in the real-world performance evaluation focusing on the aforementioned aspects, the baseline models have exhibited notable deficiencies, revealing the potential risks of using them in practical applications.
In the evaluation, the newly proposed unbiased benchmark allows these shortcomings to be identified more clearly, emphasizing the necessity for its introduction.

\noindent The contribution of this paper are summarized as follows: 
\begin{enumerate}[leftmargin=*,noitemsep,topsep=0pt]

    \item We point out that the capability of PTLMs on real-world VrD-IE cannot be adequately evaluated using the existing pipeline, since current benchmarks are not properly tailored for the evaluation purpose, and the existing pipeline is constrained by a monotonous evaluation standard.

    \item We introduce EC-FUNSD, a entity-centric dataset of SER and EL that serves as a precise benchmark to evaluate the capabilities of PTLMs. 
    This dataset has eliminated the spurious correlation bias in the previous dataset to support the real-world VrD-IE evaluation of PTLMs. 

    \item Our experiments with prevalent PTLMs show that these models may not perform as good as they claimed in real-world VrD-IE tasks, and are facing challenges related to generalization, robustness and fairness in practical applications. 

\end{enumerate}

\noindent We anticipate that this research will inspire the community to carry out more appropriate evaluation for existing methods, and develop novel methods that are more adaptable to real-world applications.

\section{Related Work}

\subsection{Pre-trained Text-and-Layout Models}

Driven by the success of contextualized pre-trained language models in diverse NLP tasks and the growing demands of document AI, extensive studies have integrated multimodal features-such as layout and image information—into pre-trained language models, leading to the emergence of PTLMs.
The effectiveness of these models in document representation has been successfully validated across various downstream VrD tasks.

The LayoutLM series \cite{xu2020layoutlm,xu2021layoutlmv2,huang2022layoutlmv3,xu2021layoutxlm} established the foundation for the prevailing paradigm of PTLMs, drawing inspiration from the common practice of representative pre-trained language models \cite{kenton2019bert,liu2019roberta,sun2019ernie}.
Building upon the BERT architecture which integrates text and position embedding within inputs, LayoutLM further incorporates token-level 2-D position embedding together with image embedding.
It is pre-trained on document layouts with masked visual-language modeling objective to achieve an effective fusion of text, layout, and visual information, significantly improving the performance on VrD tasks. 
Following the paradigm, many PTLMs have been developed, employing diverse pre-training strategies to enhance the interaction between multimodal features \cite{li2021structext,lyu2024structextv3,zhai2023fast,tu2023layoutmask,appalaraju2021docformer,appalaraju2024docformerv2,hong2022bros}.
Moreover, to better tackle specific downstream tasks, certain PTLMs have been designed to align with targeted distributions through carefully crafted pre-training strategies.
For example, \cite{DocRel-li2022relational,krishnan2024towards,peng2022ernie,cvpr2023geolayoutlm,yu2023structextv2} introduced relation-aware pre-training to better capture the spatial and semantic relationships between layout elements to enhance entity linking performance. 
Similarly, \cite{wang2022queryform,yang2023modeling,tu2024uner} introduced query-aware architecture to improve performance on open-set SER through large-scale pre-training, particularly in few-shot and zero-shot scenarios.
To address the challenges of representing and understanding long documents, \cite{yao2023resuformer,jiang2024towards} focused on modeling the the hierarchical structure of layout elements to capture both global and local features within documents. 
While these approaches demonstrate strong performance on downstream tasks, their effectiveness in real-world applications remains to be fully validated.

\begin{figure}[t]
  \includegraphics[width=0.45\textwidth]{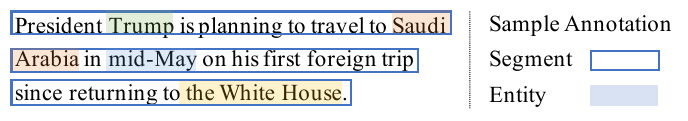}
  \vspace{-2.5mm}
  \caption{Document sample with layout and entity annotations, where the entity "Saudi Arabia" spans across segments.}
  \vspace{-4mm}
  \label{fig:entity_layout}
\end{figure}

\subsection{Generalization, Robustness, and Fairness Evaluation for Deep Learning Systems}

Considering that the focus of generalization, robustness and fairness may vary across different contexts, this section first specifies their concepts within this paper, and then introduces previous studies on the general principles and specific methods for the evaluation of these aspects.

\paragraph{Generalization}
Generalization refers to the ability of a trained model to make accurate predictions on new, unseen data. In real-world applications, the out-of-distribution (OOD) generalization issue leads to performance degradation as the data distribution in real-world applications may differ from the training distribution, highlighting the necessity of generalization evaluation. 
\citet{Hupkes2023} identified key dimensions of generalization evaluation, including compositional, structural, cross-task, cross-lingual, cross-domain, and robustness generalization. 
\citet{elangovan2024principles} introduced internal and external validity as key measurements for generalization evaluation. 
Internal validity measures the capability of methods to capture cause-and-effect relationships, while external validity pertains to task-specific generalization.  
From this perspective, the OOD generalization challenge is categorized as an external validity challenge, and an evaluation framework for cause-and-effect modeling is proposed. 

\paragraph{Robustness}
In this paper, we focus on the natural robustness of PTLMs, i.e. the robustness to natural distribution shifts.
Ideally, the model performance should remain stable when the inputs are drawn from any distribution that is naturally occurring in real-world scenarios. 
\citet{wang2021measure} summarized the concept of natural robustness with two key assumptions: label-preserving and semantic-preserving.  
The label-preserving assumption implies that human predictions should remain unchanged despite perturbations in the input.  
The semantic-preserving assumption states that perturbed inputs should retain semantic consistency with the original inputs.
Based on these assumptions, several studies have designed robustness evaluation falling into two categories: (1) those where the inputs change significantly while labels remain unchanged, and (2) those where the inputs remain relatively stable but the labels change.
\citet{wang2021textflint} introduced a unified multilingual robustness evaluation toolkit that integrates universal linguistically based text transformations, task-specific transformations, adversarial attacks, subpopulation analysis, and their combinations to provide comprehensive robustness assessment.

\paragraph{Fairness}
Our focus of fairness in this paper is on the equality of opportunity (group fairness) of PTLMs, which requires the model performance remains stable across any subgroup.   
\cite{chang-etal-2019-bias,bansal2022survey,10.1145/3616865,dong2023fairness} have pointed out that the concerns of algorithmic fairness vary across different applications, influencing the design principles of fairness enhancement methods.
According to these studies, common group fairness concepts include demographic parity, equality of odds, and equality of opportunity.
Based on these concepts, these studies identified the potential drawbacks of current deep learning systems, scenarios that would induce bias, and introduced metrics to quantify these biases. They also proposed design principles and solutions to mitigate bias and improve fairness.

\renewcommand\tabcolsep{5pt}
\begin{table}[t]
\centering
\caption{The proportion of complex entities in each benchmark dataset.}
\vspace{-2mm}
\label{tab:stats_complex_entity}
\begin{tabular}{c|c}
    \bottomrule
    \textbf{Dataset} & \textbf{The proportion of complex entities} \\
    \hline
    EPHOIE \cite{ephoie} & $0/9,823 = 0.00\%$ \\
    CORD \cite{park2019cord} & $266/13,515 = 1.96\%$ \\
    FUNSD \cite{jaume2019funsd} & $576/8,529 = 6.75\%$ \\
    RFUND \cite{lin2024peneo} & $7,098/10,4184 = 6.81\%$ \\
    SROIE \cite{huang2019icdar2019} & $925/3,780 = 24.47\%$ \\
    \hline
    EC-FUNSD & $682/8,398 = 8.12\%$ \\
    \toprule
    \end{tabular}
    \vspace{-4mm}
\end{table}

\section{EC-FUNSD: An Entity-Centric VrD-IE Benchmark Dataset}

\subsection{Defect Analysis for Prevailing VrD-IE Benchmark Datasets} 
\label{sec:dataset_motivation}

In previous works, popular datasets include FUNSD \cite{jaume2019funsd}, XFUND \cite{xu2022xfund}, CORD \cite{park2019cord}, EPHOIE \cite{ephoie}, SROIE \cite{huang2019icdar2019}, FUNSD-r and CORD-r \cite{zhang2023reading} have been adopted in the evaluation of PTLMs. 
Despite the widespread use of these datasets, they still have several limitations that hinder them from being suitable and reliable for evaluating the VrD-IE capabilities of PTLMs. 
The defect analysis of these datasets highlights the necessity of establishing new VrD-IE benchmarks. 

\definecolor{myred}{RGB}{255, 0, 0}
\definecolor{mygreen}{RGB}{0, 192, 0}
\newcommand{\mytick}{{{\color{mygreen}\ding{51}}}}
\newcommand{\mycross}{{{\color{myred}\ding{55}}}}

\renewcommand\tabcolsep{5pt}
\begin{table*}[t]
\centering
\caption{The requirements of benchmarks being suitable for evaluating the VrD-IE capabilities of PTLMs. }
\vspace{-2mm}
\label{tab:mark}
\begin{tabular}{c|c|ccccc|c}
    \bottomrule
    \textbf{Dimension} & \textbf{Requirement} & FUNSD & SROIE & CORD & EPHOIE & FUNSD-r & EC-FUNSD \\
    \hline
    \multirow{3}{*}{\makecell[c]{Layout Annotation\\Quality}} & Segment- and word/char-level annotations & \mytick & \mycross & \mytick & \mytick & \mytick & \mytick \\
     & Same distribution to OCR engine & \mycross & \mytick & \mycross & \mycross & \mytick & \mytick \\
     & High annotation quality & \mytick & \mytick & \mycross & \mytick & \mycross & \mytick \\
    \hline
    \multirow{3}{*}{\makecell[c]{Entity Annotation\\Quality}} & Decoupled from layout annotation & \mycross & \mytick & \mycross & \mytick & \mytick & \mytick \\
     & Annotated in word/char-level & \mycross & \mycross & \mycross & \mytick & \mytick & \mytick \\
     & Diverse and complex entities & \mytick & \mytick & \mycross & \mycross & \mytick & \mytick \\
    \hline
    Suitable to PTLM & Following sequence-labeling paradigm & \mytick & \mytick & \mytick & \mytick & \mycross & \mytick \\
    \toprule
    \end{tabular}
    \vspace{-3mm}
\end{table*}

The most influential benchmarks are not well-suited to the VrD-IE evaluation due to their specific limitations. 
For FUNSD and XFUND, their annotation does not conform to the SER task settings.
Specifically, these datasets are adapted from the visual information extraction task, and their annotations are organized by visual regions (i.e., blocks or segments) of the document layout, with the category labels and association relationships annotated on the block-level. 
However, the bounding box annotation of these blocks does not take the semantic of contents into account, resulting in massive cases where the contents within a block may not correspond to an entity.
In real-world scenarios, document layout annotations are usually generated by an off-the-shelf OCR engine, which focuses solely on visual features while ignoring semantic features. 
As a result, there are cases that a block contain multiple entities, or an entity spans across multiple blocks, as shown in Fig. \ref{fig:entity_layout}.
Nevertheless, current evaluation pipeline would use the block-level annotations to represent both segments and entities in SER and EL tasks, which introduces ambiguity and interferes with accurate performance evaluation.
As shown in Fig. \ref{fig:page1}(1), the paragraph is split into two blocks, both labeled as "answer" and linked to the preceding "question" block, yet the proper annotation should treat the whole paragraph as a single semantic entity as the "answer" to the preceding "question". 
Due to this phenomenon, FUNSD and XFUND are not suitable for evaluation, as well as \cite{yang2023modeling,sun2021spatial-wildreceipt,vsimsa2023docile} which share the same issue.  
\cite{huang2019icdar2019,lin2024peneo,yu2023icdar} have tackled this issue by decoupling the annotation of blocks and entities.
However, these datasets lack of word- or char-level layout annotations, and their entity annotations are still at block-level. According to Fig. \ref{fig:entity_layout}, clearly indicating entities requires word-level annotations. 
Thus, these datasets are still unsuitable for reliably evaluating the VrD-IE capabilities of PTLMs. 
CORD and EPHOIE suffer from a lack of diversified layout formats and semantic entities. 
Most samples in CORD are receipts with similarly simple layouts, and the entities they contain are highly repetitive across samples -- many samples even share identical entity content.
Most samples in EPHOIE are long, narrow information columns from test papers with simple layouts and sparse text. Since EPHOIE was originally designed for key-value extraction tasks, each type of entity appears at most once in each sample, limiting the interactions between entities of the same type.
Besides, as shown in Tab. \ref{tab:stats_complex_entity}, complex entities, such as those span multiple rows/columns or are interrupted by other contents, are lacking in these datasets. 
However, accessing the model's capability of recognizing such challenging entities is crucial for evaluation, indicating that CORD and EPHOIE are unsuitable for evaluating the VrD-IE capabilities of PTLMs.
FUNSD-r and CORD-r acknowledge the importance of fine-grained annotations, offering decoupled char-level layout and entity annotations. 
However, these datasets adopted a special schema for entity annotation, which do not guarantee the continuity of entities contained in the word sequence. 
Therefore, these datasets cannot be completely tackled by sequence-labeling models and cannot directly be used to evaluate PTLMs. 
Besides, their layout annotations are generated by automated OCR engines and are of low quality. 

To sum up, current benchmarks fall short of meeting the requirements for the evaluation.
In this paper, our aim is to identify the problems of using existing PTLMs in real-world VrD-IE applications with quantifying the negative impacts. 
Thus, it is essential to propose new evaluation pipeline with new benchmarks to enable reliable and comprehensive evaluation. 

\renewcommand\tabcolsep{5pt}
\begin{table*}[t]
\begin{spacing}{1.19}
\begin{adjustwidth}{-.0in}{-.0in}
\centering
\small
\caption{Statistics of the proposed dataset. \# is short for "Number of". The statistics of FUNSD is also listed in comparison, with invalid entity linking pairs removed.}
\vspace{-2mm}
\label{tab:stats}
\begin{tabular}{c|ccccc}
    \bottomrule
    \textbf{Dataset} & 
    \bfseries\makecell[c]{\# Segments} & 
    \bfseries\makecell[c]{\# Words} & 
    \bfseries\makecell[c]{\# Segs. per Sample} &
    \bfseries\makecell[c]{\# Words per Sample} &
    \bfseries\makecell[c]{Avg. Len. of Segment} \\
    \hline
    \textbf{FUNSD} & 9,743 & 31,485 & 48.95 & 158.21 & 3.23 \\
    \textbf{EC-FUNSD} & 10,662 & 31,297 & 53.57 & 157.27 & 2.93 \\
    \Xhline{2\arrayrulewidth}
    \textbf{Dataset} & 
    \bfseries\makecell[c]{\# Entities} & 
    \bfseries\makecell[c]{\# Ents. per Sample} &
    \bfseries\makecell[c]{Avg. Len. of Entity} & 
    \bfseries\makecell[c]{\# Relation Triplets} &
    \bfseries\makecell[c]{\# Rel. Triplets per Sample} \\
    \hline
    \textbf{FUNSD} & 8,529 & 42.85 & 2.92 & 3,966 & 19.92 \\
    \textbf{EC-FUNSD} & 8,398 & 42.20 & 2.96 & 3,912 & 19.65 \\
    \toprule
\end{tabular}
\end{adjustwidth}
\end{spacing}
\vspace{-3mm}
\end{table*}

\subsection{Construction of EC-FUNSD}

Based on the review of prevailing datasets, we claim that evaluating PTLMs with these benchmarks may be imprecise to reflect their real VrD-IE capabilities. 
In order to establish a high-quality dataset that is suitable to the evaluation, we propose five essential requirements for an appropriate benchmark: 
(1) The layout annotations should represent the typical outputs generated by OCR engines, including segment-level text regions (as text lines) and fine-grained character (or word) bounding boxes; 
(2) The layout annotations should be of high quality, with minimal omission of words and segments; 
(3) The annotation of semantic entities should confirm to unified semantic-driven definitions to ensure consistency across samples; 
(4) Each sample should be a single-page document that contains rich layout and diversified entities and relations to increase the validness of evaluation on the benchmark; 
and (5) The layout annotation of the text of semantic entities needs to be continuous to meet the requirements for the sequence-labeling paradigm in the SER task.

To construct a dataset that satisfies these requirements and serves as a proper benchmark for the evaluation, we choose to revise the existing annotations within the FUNSD dataset due to the following reasons: (1) Compared with other prevalent benchmarks, FUNSD stands out by offering a diverse range of layouts among its samples, rendering it a comprehensive benchmark. (2) FUNSD contains extensive block annotations with various semantic types and their relations. These existing annotations can be revised into semantic entity and relation annotations with little modification.

Therefore, based on the original dataset, we executed a two-step revision of the layout and entity annotations to create a \textbf{E}ntity-\textbf{C}entric version of \textbf{FUNSD}, namely \textbf{EC-FUNSD}. 
In the first step, we constructed the word and segment-level layout annotations by manually revising the original layout annotations of each sample. We removed empty words in the original layout annotations, and marked unrecognizable handwritten words as "<unk>". We appended omitted words, associating them to appropriate existing segments or creating new segments when necessary. We rectified all errors on texts or bounding boxes of words and segments. We removed words that are of low resolution and deemed unimportant to the remaining contents, because their original annotations are erroneous and we are unable to correct these illegible words. We manually split multiple-row blocks within the original annotations into multiple segments, each segment confined in one row. After row-splitting, we combined segments that are tightly adjacent to each other into one segment. Throughout this process, we preserved the sequential order of words within each segment and also the mapping of words and segments from revised annotations to the old ones, to ensure the mapping of entity and relation annotations is preserved.
After correcting layout annotations, in the second step, we revised the entity and relation annotations by manually transforming the block annotations to semantic entity annotations. We mapped the original block annotations to form the preliminary entity annotations for revision. After that, we corrected the annotation of entities being annotated across multiple semantic blocks, and entities with missing words in their annotations. Additionally, we removed the invalid linking pairs and modified the corresponding linking pairs following the modifications made to the entity annotations. We generally preserved the segment order within the original annotations to ensure that each entity span is continuous in the annotations, making the form of this dataset suitable for sequence-labeling models. The annotating procedures above are carried out by two qualified annotators who are familiar with document AI.

\subsection{Statistics of EC-FUNSD}

The statistics of EC-FUNSD are displayed in Tab. \ref{tab:stats}. EC-FUNSD has a smaller total word count than FUNSD, primarily due to the removal of low-quality word annotations; The number of segments increases since multiple-row blocks are split to several segments; The average length of entities slightly increases as the entities that were split into multiple semantic blocks were combined.

Compared with the aforementioned benchmarks, these key factors of EC-FUNSD making it well-suited for the evaluation of the real-world VrD-IE capabilities of PTLMs:
(1) EC-FUNSD offers high-quality annotations. These include word- and segment-level layout annotations, along with manually-corrected entities and linking relationships, laying a solid foundation for evaluation.
(2) The entity annotations in EC-FUNSD follow the sequence labeling paradigm, facilitating the evaluation.
(3) EC-FUNSD features diverse layout patterns, mirroring real-world document layouts. It also includes a sufficient number of complex entities, enabling comprehensive evaluation of model generalization, robustness, and fairness in real-world scenarios. 

\section{Real-world VrD-IE Evaluation for Pre-trained Text-and-Layout Models}

Based on the proposed dataset, we are now able to conduct a comprehensive evaluation of the capabilities of PTLMs in real-world VrD-IE. 
In this section, we first present the specific VrD-IE task settings we focus on, followed by the motivation and implementation of our evaluation methods.

\subsection{VrD-IE Task Formulation}
\label{sec:formulation}

The SER and EL tasks on document layouts are formalized as follows. 
A document layout with $N_{\mathcal{D}}$ words is represented as $\mathcal{D}=\{(w_i, \textbf{b}_i)\}_{i=1,\dots,N_{\mathcal{D}}}$, where $w_i$ denotes the $i$-th word in document and $\textbf{b}_i=(x_i^0, y_i^0, x_i^1, y_i^1)$ denotes the position of $w_i$ in the document layout. The coordinates $(x_i^0, y_i^0)$ and $(x_i^1, y_i^1)$ correspond to the bottom-left and top-right vertex of $w_i$'s bounding box, respectively. 
Given the predefined semantic entity types $\mathcal{E}=\{e_i\}_{i=1,\dots,N_{\mathcal{E}}}$ and relation types $\mathcal{R}=\{r_i\}_{i=1,\dots,N_{\mathcal{R}}}$, 
the semantic entities within document $\mathcal{D}$ is denoted as $s_{\mathcal{D}} = \{s_1,\dots,s_J\}$, where the $j$-th entity $s_j=\{e_j,(j_1,j_2)\}$ is identified by its entity type $e_j \in \mathcal{E}$ and an index span $(j_1,j_2)$ indicating the position of words in the inputs, satisfying $1 \leq j_1 \leq j_2 \leq N_{\mathcal{D}}$. 
The set of relationships between entities of $s_{\mathcal{D}}$ is denoted as $t_{\mathcal{D}} = \{t_1,\dots,t_K\}$, where the $k$-th relation triplet $t_k=\{r_k,(ks,ko)\}$ indicates that the relation between the subject $s_{ks}$ and object $s_{ko}$ is $r_k \in \mathcal{R}$ ($1 \leq ks, ko \leq J$). 
It is guaranteed that there would be at most one relation triplet from one entity to another.
The aim of SER is to recognize all the entity that spans together with their semantic categories, while EL aims to identify the possible relationship between two arbitrary entities, as well as type of relationship in the document. 

\begin{figure*}[htbp]
  \includegraphics[width=\textwidth]{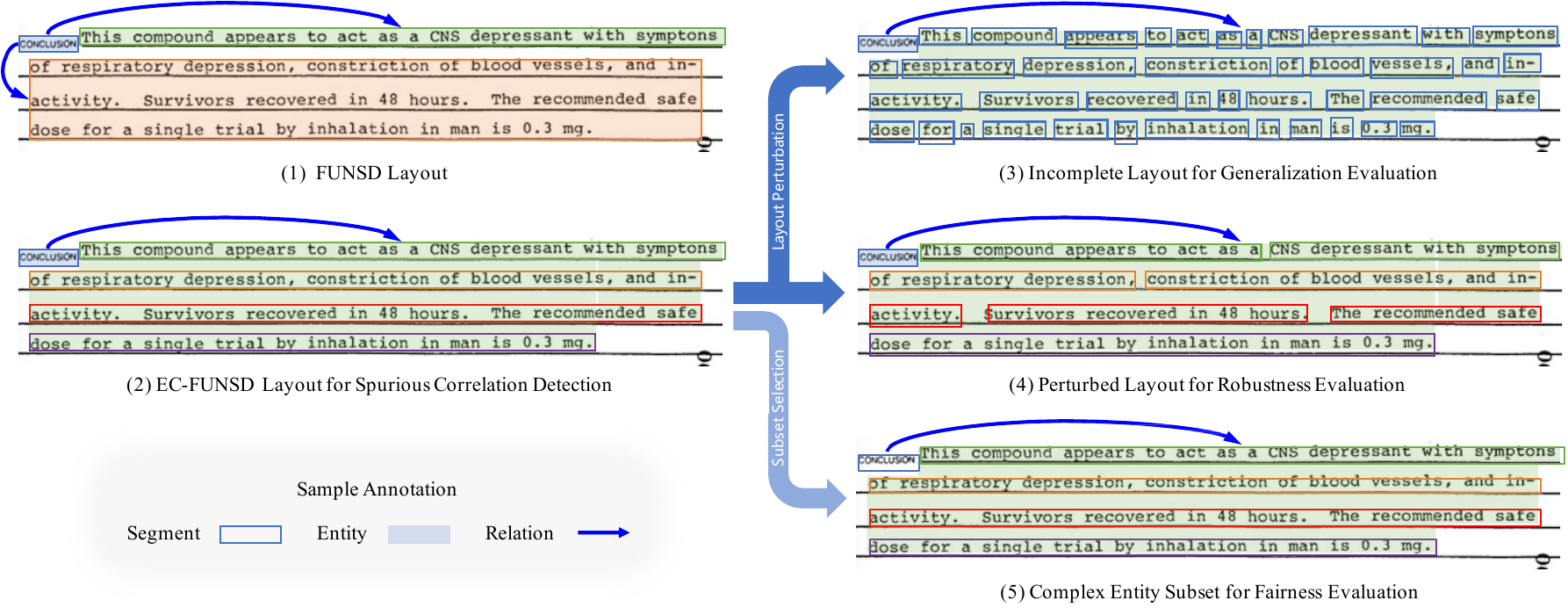}
  \vspace{-6mm}
  \caption{Document layout samples used in each evaluation.}
  \vspace{-1mm}
  \label{fig:main}
\end{figure*}

\subsection{Spurious Correlation Detection}

This evaluation aims to identify potential spurious correlations between input features and output labels in the prevailing datasets and quantify its influence to model evaluation. 
As mentioned in \S\ref{sec:dataset_motivation}, when evaluating on SER with FUNSD-like datasets \cite{jaume2019funsd,xu2022xfund,yang2023modeling,sun2021spatial-wildreceipt,vsimsa2023docile}, the block-level annotations simultaneously represent both segments and entities. 
Therefore, models would predict entity boundaries and types specified by block annotations while leveraging the segment-level layout inputs from the same block annotations, which leads to a potential risk that models would overly rely on correlations between input layout features and entity labels.
During training and testing, input units (e.g., tokens in a Transformer model) within the same block share the same layout feature (the block's bounding box), which leaks the block boundary information that need to be predicted. 
Therefore, models quickly converge during training by fitting these identical layout input features.
However, the correlation between input layout features and entity labels is spurious and specific to certain datasets.
In real-world SER, there is typically no inherent strong link between visual-based semantic block layout information and semantic-based entity boundaries. 
Thus, this correlation should not be leveraged as a reliable task feature. 

In this evaluation, we measure the real performance of PTLMs on VrD-IE by removing the spurious input-label correlations. 
We apply PTLMs on EC-FUNSD for SER and EL tasks and compare the results with FUNSD, to determine whether current PTLMs overfit the spurious input-label correlations and quantify the corresponding negative impact.

\subsection{Generalization Evaluation}

The purpose of this evaluation is to examine the generalization ability of PTLMs to the distribution shift of real-world datasets. 
Different from standardized benchmark datasets that completely conform to the task formulation (\S\ref{sec:formulation}), real-world task datasets may exhibit incompleteness of layout features.
Existing PTLMs are designed for and pre-trained on structured documents (contracts, forms, etc.) and leverage both char/word- and segment-level layouts as input features. 
However, the processing demands in real-world scenarios often extend beyond standardized textual formats, encompassing more diverse document types such as invoices, tickets, cards, WordArts, and UI screenshots.
The layout features of these documents are usually auto-extracted from OCR engines, yielding only char/word-level layout information, while segment-level layouts are often absent, or too noisy to be effectively utilized.
For example, being the most popular benchmark for document reading order prediction, ReadingBank \cite{wang2021layoutreader} only offers word-level layout annotations;
typical Chinese VrD benchmarks, such as CTW \cite{yuan2018chinese-ctw}, provide only char-level layout annotations;
and synthesized datasets like EATEN \cite{guo2019eaten, kim2022ocr-donut} also rely solely on single-level auto-generated layout annotations.
Therefore, in generalization evaluation, we simulate the distribution shift in real-world scenarios by masking segment-level layout information and keeping only word-level layout information as input, evaluating the generalization ability of PTLMs by measuring their performance changes when adapting to the shifted distribution.

\subsection{Robustness Evaluation}

This evaluation is designed to assess the robustness of PTLMs against the real-world input perturbations. 
The processing of real-world documents may suffer from the printing distortion with overlapped, blurred or misaligned texts. 
Scanning-related issues like skew, jitter, or inconsistent lighting can further degrade the quality of scanned document pages, negatively influencing the model performance. 
In this evaluation, we apply random shifting and rotation to the input bounding boxes to simulate real-world perturbations, measuring the extent to which current PTLMs are affected by real-world layout perturbations.

\subsection{Fairness Evaluation}

The aim of this evaluation is to test the consistency of model performance on specific subset of samples. 
In real-world applications, we expect that the model performance on any subset of samples should not be significantly lower than its average performance, ensuring the consistency and reliability across diverse data distributions that the model remains robust and fair even in less frequent or challenging cases.
In tackling SER task with PTLMs, we focus on complex entities that span multiple rows and columns and overlap with other entities in region.
As shown in Fig. \ref{fig:page1}(2), the entire paragraph corresponds to a single entity, whose complex layout patterns like line wrapping and indentation making it extremely difficult for the model to accurately predict the boundaries of it. 
Additionally, due to the nature of the sequence-labeling paradigm, false prediction on any single token within the long entity can result in the failure to correctly recognize it. 
Therefore, this evaluation focuses on the capability of PTLMs on recognizing complex entities, and we take it as a key dimension for measuring the real-world fairness of PTLMs.

\section{Experiments}

\subsection{Implementation Details}

We use two baseline PTLMs to be evaluated: LayoutLMv3 \citep{huang2022layoutlmv3} as the most popular PTLM, and GeoLayoutLM \citep{cvpr2023geolayoutlm} as the current state-of-the-art PTLM. 
LayoutLMv3 enhances the perception of visual signals by pre-training to align the inputs of modalities. 
GeoLayoutLM introduces multi-level geometric pre-training tasks to model spacial relationships between layout elements. 


We fine-tune the two baseline models for SER and EL tasks on FUNSD and EC-FUNSD, using F1 score metrics for spurious correlations evaluation, generalization evaluation and robustness evaluation while using the recall for fairness evaluation. 
In spurious correlation detection, we fine-tune models in their normal way. 
In generalization evaluation, we use word-level layout as input instead of segment-level layout to simulate the distribution offset of real-world datasets. 
We measure the generalization ability of PTLMs on two settings, namely real-world generalization, where the distribution shifts of inputs are applied only during inference; and real-world adaptation, where the shifts are applied during both training and inference.
In robustness evaluation, the layout input is perturbed to simulate the random deviation in the real-world scanning results. 
Specifically, we first simulate the falsely split text regions by sampling cutoff length with $Exp(0.1)$ distribution and splitting each segment with the sampled value. 
Next, each text region is centrally rotated $U(-5, 5)$ degrees to simulate rotational misalignment during scanning. 
Finally, an offset of $N(0, U(5, 20)^2)$ pixels is added to the upper, lower, left and right boundaries of each text region to simulate the deviation in the recognition result. 
Similar to generalization evaluation, we also measure the robustness of PTLMs on two settings, namely real-world robustness where the perturbations of inputs are applied only during inference, and adversarial training (AT) where the perturbations are applied during both training and inference.
In the fairness experiment, we also fine-tune the models in the normal way, but evaluate them with their recall on the whole set of entities and subset of complex entities.

The detailed configuration of fine-tuning these models are further illustrated as follows.
In experiments, we use the official implementation and pre-trained weights of LayoutLMv3-base \footnote{\url{https://github.com/microsoft/unilm/tree/master/layoutlmv3}} \citep{huang2022layoutlmv3} and GeoLayoutLM \footnote{\url{https://github.com/AlibabaResearch/AdvancedLiterateMachinery/tree/main/DocumentUnderstanding/GeoLayoutLM}} provided by their official GitHub repositories. 
It is important to note that the predefined maximum sequence length of textual tokens in both models is limited to 512. Therefore, when processing long documents that surpass this limit, LayoutLMv3 divides the document into several segments, whereas GeoLayoutLM truncates the content beyond the maximum length.  
Both means inevitably disrupt the integrity of EL labels in the documents, resulting in unfair comparison with other methods. 
To address this issue, we increase the maximum sequence length to 1024 by initializing the positional embedding of index 512-1023 by those of index 0-511 before fine-tuning. This adjustment guarantees none of the training or validation samples exceed the maximum length, and results in a slight difference compared to the results proposed in the original releases of the baselines.

The hyperparameters of fine-tuned the baseline models on FUNSD and EC-FUNSD for SER and EL tasks are reported as follows.
In fine-tuning LayoutLMv3 for SER, we follow all the original setting of \citep{huang2022layoutlmv3}. 
In fine-tuning LayoutLMv3 for EL, we generally follow all the original setting with 400 epochs of fine-tuning. 
In fine-tuning GeoLayoutLM for SER and EL, for better performance, instead of following the original settings proposed in \citep{huang2022layoutlmv3}, we use an AdamW optimizer with 2\% linear warming-up steps and a 1e-2 weight decay with a cosine scheduler. 
The learning rate and batch size were 1e-5 and 16 as the optimal configure searching from lr=\{8e-6, 1e-5, 1.5e-5, 2e-5\} and bs=\{6, 16\}. 
All models are fine-tuned by 500 epochs and the checkpoints with the best performance on SER and EL are kept.
Generally, we ensure the consistency in fine-tuning and evaluating on FUNSD and EC-FUNSD, with the sole exception that we disabled the vision branch of GeoLayoutLM when fine-tuning on EC-FUNSD. 
In specific, we notice that the vision feature was only available in block-level in GeoLayoutLM, which directly contributes to the entity feature when fine-tuning FUNSD. 
However, a corresponding block-level vision representation is not available for every entity in EC-FUNSD, e.g. for the entities that span multiple rows and overlap with other entities in region. 
Therefore, the vision inputs are disabled in fine-tuning GeoLayoutLM on EC-FUNSD.

\begin{table}[t]
    \centering
    \caption{Performance of baseline models on FUNSD and EC-FUNSD.}
    \vspace{-2mm}
    \label{tab:spurious}
    \begin{tabular}{c|c|c|c}
        \bottomrule
        \textbf{Task} & \textbf{Model} & \textbf{FUNSD} & \textbf{EC-FUNSD} \\
        \hline
        \multirow{3}{*}{SER} 
        & LayoutLMv3-base & 90.85 & 82.30 ($\downarrow$8.55) \\
        & LayoutLMv3-large & 91.70 & 83.88 ($\downarrow$7.82) \\
        & GeoLayoutLM & 91.10 & 83.62 ($\downarrow$7.48) \\
        \hline
        \multirow{3}{*}{EL} 
        & LayoutLMv3-base & 69.80 & 67.47 ($\downarrow$2.33) \\
        & LayoutLMv3-large & 79.37 & 78.14 ($\downarrow$1.23) \\
        & GeoLayoutLM & 88.06 & 86.18 ($\downarrow$1.88) \\
        \toprule
    \end{tabular}
    \vspace{-2.5mm}
\end{table}

\subsection{Analysis on Spurious Correlation of FUNSD}

Tab. \ref{tab:spurious} shows the standard fine-tuning results of baseline models on FUNSD and EC-FUNSD. 
Although EC-FUNSD and FUNSD are very similar in almost all aspects, three models perform significantly worse on EC-FUNSD than on FUNSD, especially on the SER task, with a 7.48-8.55 drop in F1 scores on EC-FUNSD. We attribute the performance degradation to false overfitting on the block-level text region layout feature for training and inference. As illustrated
in \S\ref{sec:dataset_motivation}, FUNSD derives its segment and entity annotations directly from blocks, which leads to considerable bias when fine-tuning on this dataset. For SER, tokens within the same entity have exactly the same xy-coordinate inputs, inducing the models to learn entity boundaries simply by determining whether the layout features between tokens are consistent or not. For EL, the entity representations are heavily dominated by the layout features, as all tokens within an entity share the same layout features, resulting in less attention to entity semantics.
This experiment reveals the potential risk of current PTLMs that they may develop to excessively overfit the biased benchmarks, but the actual benefits brought by the advancements are suspicious. 

Additionally, according to the experiment, EC-FUNSD serves as a better indicator of model performance than FUNSD. 
The performance gap between FUNSD and EC-FUNSD illustrates that the bias in FUNSD significantly influences its trustworthiness as an evaluation benchmark. 
In evaluation on FUNSD, since spurious correlations exist in both the training and validation sets, this issue is not reflected in the performance metrics. 
In contrast, the evaluation on EC-FUNSD effectively reveals the shortcomings: once the spurious correlations in the dataset are removed, the model performance would drop sharply.

\subsection{Analysis on Generalization of PTLMs}

\renewcommand\tabcolsep{5pt}
\begin{table}[t]
    \begin{spacing}{1}
        \begin{adjustwidth}{-.0in}{-.0in}
            \centering
            \small
            \caption{Performance of baseline models in generalization evaluation.}
            \vspace{-2mm}
            \label{tab:generalization}
            \subcaption{Performance on FUNSD.}
            \begin{tabular}{c|c|c|c|c}
                \bottomrule
                \textbf{Task} & \textbf{Model} & \textbf{Ori.} & \textbf{Generalization} & \textbf{Adaptation} \\
                \hline
                \multirow{3}{*}{SER} 
                & LayoutLMv3-base & 90.85 & 22.75 ($\downarrow$68.10) & 81.06 ($\downarrow$9.79) \\
                & LayoutLMv3-large & 91.70 & 19.92 ($\downarrow$71.78) & 83.23 ($\downarrow$8.47) \\
                & GeoLayoutLM & 91.10 & 19.47 ($\downarrow$71.63) & 81.70 ($\downarrow$9.40) \\
                \hline
                \multirow{3}{*}{EL} 
                & LayoutLMv3-base & 69.80 & 53.83 ($\downarrow$15.97) & 63.29 ($\downarrow$6.51) \\
                & LayoutLMv3-large & 79.37 & 69.43 ($\downarrow$9.94) & 77.92 ($\downarrow$1.45) \\
                & GeoLayoutLM & 88.06 & 59.66 ($\downarrow$28.40) & 84.98 ($\downarrow$3.08) \\
                \toprule
            \end{tabular}
            \vspace{1mm}
            \subcaption{Performance on EC-FUNSD.}
            \begin{tabular}{c|c|c|c|c}
                \bottomrule
                \textbf{Task} & \textbf{Model} & \textbf{Ori.} & \textbf{Generalization} & \textbf{Adaptation} \\
                \hline
                \multirow{3}{*}{SER} 
                & LayoutLMv3-base & 82.30 & 50.53 ($\downarrow$31.77) & 79.74 ($\downarrow$2.56) \\
                & LayoutLMv3-large & 83.88 & 42.48 ($\downarrow$41.40) & 83.25 ($\downarrow$0.63) \\
                & GeoLayoutLM & 83.62 & 38.04 ($\downarrow$45.58) & 82.38 ($\downarrow$1.24) \\
                \hline
                \multirow{3}{*}{EL} 
                & LayoutLMv3-base & 67.47 & 61.97 ($\downarrow$5.50) & 64.80 ($\downarrow$2.67) \\
                & LayoutLMv3-large & 78.14 & 72.47 ($\downarrow$5.67) & 77.34 ($\downarrow$0.80) \\
                & GeoLayoutLM & 86.18 & 78.11 ($\downarrow$8.07) & 84.66 ($\downarrow$1.52) \\
                \toprule
            \end{tabular}
        \end{adjustwidth}
    \end{spacing}
    \vspace{-4.5mm}
\end{table}  

Tab. \ref{tab:generalization} presents the generalization evaluation results of baseline PTLMs. 
Real-world generalization simulates scenarios where a deployed PTLM must handle input samples lacking segment-level layout features, thereby evaluating the ability of PTLMs to generalize to unseen input distributions.
Real-world adaptation mimics scenarios where PTLMs must adjust to specific downstream tasks that lack segment-level features, evaluating the adaptability of PTLMs to real-world tasks with distributions differing from those seen during pre-training and fine-tuning. 

According to the results, 
(1) Despite achieving strong performance on prevailing benchmarks, current PTLMs fall short in real-world scenarios.
In generalization adaptation, the performance of PTLMs decline on both datasets. Notably, for the SER task on FUNSD, the F1 scores of three models drop significantly by 8-10\%. 
(2) EC-FUNSD exhibits its superiority as an evaluation benchmark. 
In generalization adaptation, the performance of GeoLayoutLM drops significantly on FUNSD by over 9\% and 3\% on SER and EL, despite using both word- and segment-level layout features.
We hypothesis that GeoLayoutLM overly relies on segment-level layout features after fine-tuning on FUNSD, while failing to fully leverage word-level layout and semantic features. 
This further highlights the negative impact of the spurious correlations in FUNSD for evaluation, casting doubt on the previous results. 
In contrast, the results on EC-FUNSD exhibit smaller drops under both settings, indicating that PTLMs trained on EC-FUNSD rely less on segment-level layout features. 
The results also verify that the raw performance of PTLMs on EC-FUNSD is not inflated by spurious correlations, and thus can serve as a more credible metric to reflect the real-world VrD-IE capabilities of PTLMs.

\subsection{Analysis on Robustness of PTLMs}

\renewcommand\tabcolsep{5pt}
\begin{table}[t]
    \begin{spacing}{1}
        \begin{adjustwidth}{-.0in}{-.0in}
            \centering
            \small
            \caption{Performance of baseline models in robustness evaluation.}
            \vspace{-2mm}
            \label{tab:robustness}
            \subcaption{Performance on FUNSD.}
            \begin{tabular}{c|c|c|c|c}
                \bottomrule
                \textbf{Task} & \textbf{Model} & \textbf{Ori.} & \textbf{Robustness} & \textbf{AT} \\ 
                \hline
                \multirow{3}{*}{SER} 
                & LayoutLMv3-base & 90.85 & 61.41 ($\downarrow$29.44) & 83.22 ($\downarrow$7.63) \\
                & LayoutLMv3-large & 91.70 & 60.67 ($\downarrow$31.03) & 85.58 ($\downarrow$6.12) \\
                & GeoLayoutLM & 91.10 & 90.12 ($\downarrow$0.98) & 90.53 ($\downarrow$0.57) \\
                \hline
                \multirow{3}{*}{EL} 
                & LayoutLMv3-base & 69.80 & 54.13 ($\downarrow$15.67) & 62.02 ($\downarrow$7.78) \\
                & LayoutLMv3-large & 79.37 & 68.56 ($\downarrow$10.81) & 72.25 ($\downarrow$7.12) \\
                & GeoLayoutLM & 88.06 & 87.13 ($\downarrow$0.93) & 86.07 ($\downarrow$1.99) \\
                \toprule
            \end{tabular}
            \vspace{2.5mm}
            \subcaption{Performance on EC-FUNSD.}
            \begin{tabular}{c|c|c|c|c}
                \bottomrule
                \textbf{Task} & \textbf{Model} & \textbf{Ori.} & \textbf{Robustness} & \textbf{AT} \\ 
                \hline
                \multirow{3}{*}{SER} 
                & LayoutLMv3-base & 82.30 & 64.03 ($\downarrow$18.27) & 77.84 ($\downarrow$4.46) \\
                & LayoutLMv3-large & 83.88 & 69.36 ($\downarrow$14.52) & 81.19 ($\downarrow$2.69) \\
                & GeoLayoutLM & 83.62 & 82.93 ($\downarrow$0.69) & 82.75 ($\downarrow$0.87) \\
                \hline
                \multirow{3}{*}{EL} 
                & LayoutLMv3-base & 67.47 & 60.29 ($\downarrow$7.18) & 61.39 ($\downarrow$6.08) \\
                & LayoutLMv3-large & 78.14 & 71.77 ($\downarrow$6.37) & 72.35 ($\downarrow$5.79) \\
                & GeoLayoutLM & 86.18 & 85.80 ($\downarrow$0.38) & 84.12 ($\downarrow$2.06) \\
                \toprule
            \end{tabular}
        \end{adjustwidth}
    \end{spacing}
    \vspace{-0mm}
\end{table}

Tab. \ref{tab:robustness} presents the performance of baseline models in robustness evaluation. 
In real-world robustness evaluation, layout inputs are randomly perturbed during inference to simulate naturally distorted samples, mimicking real-world challenging scenarios. 
Adversarial training perturbs layout features during both training and inference to evaluate the adaptability of PTLMs in real-world applications, where layout perturbations would be commonly encountered. 

The result shows that LayoutLMv3 is significantly affected in both settings, while GeoLayoutLM exhibits little change. 
The robust performance of GeoLayoutLM can be attribute to its noise-tolerant pre-training, which introduces noise into the input layout features through the Poisson Line Segmentation algorithm -- closely resembling our perturbation settings.
Another notable observation is that GeoLayoutLM in EL task consistently performs better under robustness evaluation than under adversarial training.
Based on the results, we hypothesize that GeoLayoutLM has learned to prioritize semantic features while remaining robust to noise in layout features during pre-training.
When handling the EL task which relies more on semantic features, GeoLayoutLM is not significantly affected by layout perturbations in robustness evaluation. However, during adversarial training, the model may overfit the perturbed, low-quality training data, resulting in performance degradation.
On the other hand, models achieving better original performance on EC-FUNSD also exhibit less degradation in robustness evaluation. Additionally, the extent of performance drop on EC-FUNSD is consistently smaller than that on FUNSD. 
These two observations suggest that EC-FUNSD serves as a clear and robust indicator for evaluating model performance.

\subsection{Analysis on Fairness of PTLMs}

\begin{table}[t]
    \centering
    \vspace{2mm}
    \caption{Performance of baseline models in fairness evaluation.
        ``Total.'' and ``Comp.'' denote the recall of all entities and the subset of complex entities, respectively.}
    \label{tab:fairness}
    \resizebox{0.48\textwidth}{!}{
    \begin{tabular}{c|c c|c c}
        \bottomrule
        \multirow{2}{*}{\textbf{Model}} & \multicolumn{2}{c|}{\textbf{FUNSD}} & \multicolumn{2}{c}{\textbf{EC-FUNSD}} \\
        \cline{2-5}
        & Total. & Comp. & Total. & Comp. \\
        \hline
        LayoutLMv3-base & 91.18 & 83.97 ($\downarrow$7.21) & 84.05 & 53.29 ($\downarrow$30.76) \\
        \hline
        LayoutLMv3-large & 93.76 & 89.10 ($\downarrow$4.66) & 86.65 & 59.53 ($\downarrow$27.12) \\
        \hline
        GeoLayoutLM & 93.31 & 88.46 ($\downarrow$4.85) & 86.84 & 58.79 ($\downarrow$28.05) \\
        \toprule
    \end{tabular}
    }
    \vspace{-4mm}
\end{table}

Tab. \ref{tab:fairness} shows the performance of baseline models in fairness evaluation, from which we conclude that
(1) The fairness issue cannot be overlooked, as the recall of complex entities is notably lower than the total recall among all baseline models. 
(2) The negative impact of spurious correlations in FUNSD on evaluation is also revealed in the fairness evaluation, whereas EC-FUNSD successfully addresses this issue, highlighting its advantages for evaluation.
Due to the spurious correlations, the performance of PTLMs on the complex entity subset of FUNSD does not decline significantly, failing to expose the fairness issue. 
In contrast, the performance on EC-FUNSD drop by nearly 30\%, clearly revealing the deficiencies of PTLMs in recognizing complex entities.  

\section{Conclusion}

In this paper, our aim is to properly evaluate the capabilities of PTLMs to address real-world VrD-IE tasks. 
We question the actual performance of prevalent PTLMs in real-world VrD-IE tasks, even though they have achieved excellent performance on prevailing VrD-IE benchmarks. 
To establish a reliable evaluation pipeline, we begin by analyzing the prevailing VrD-IE benchmarks, demonstrating that none of them are suitable for the intended evaluation purposes due to their intrinsic drawbacks.
Due to the issue, we propose EC-FUNSD, an entity-centric dataset tailored for the evaluation of the real-world VrD-IE capabilities of PTLMs.
Building upon this dataset, we establish a comprehensive evaluation for PTLMs featuring multiple aspects.
Besides from their absolute performance on tasks, the evaluation also focuses on their generalization to real-world applicative tasks, robustness to real-world input perturbations, and fairness to challenging subsets.
From the analysis of extensive experiments, we draw two key conclusions:
(1) Spurious correlations between input features and labels exist in prevalent benchmarks, drawing negative impact to the evaluation.
In particular, the strong performance achieved by previous models on these benchmarks may be unreliable and fail to reflect their true capabilities.
The proposed EC-FUNSD dataset addresses the issues in previous benchmarks, and is proven suited to enable a fair and accurate evaluation.
(2) Prevalent PTLMs tend to suffer from generalization, robustness, and fairness issues, hindering their effectiveness in real-world VrD-IE applications.
We anticipate our work would suggest improvement directions for future models, and inspire future works to establish better evaluation pipelines to advance document AI.


\bibliographystyle{ACM-Reference-Format}
\bibliography{custom}


\end{document}